\documentclass[runningheads]{llncs}
\usepackage{graphicx}
\usepackage{xcolor}

\usepackage{textcomp}
\usepackage{graphicx}
\usepackage{amsmath,amssymb}
\usepackage{xspace}

\newcommand*{\ie}{i.e.\@\xspace}
\newcommand*{\cf}{c.f.\@\xspace}

\usepackage{array}
\newcolumntype{R}[1]{>{\raggedleft\let\newline\\\arraybackslash\hspace{0pt}}m{#1}}
\usepackage{arydshln}
\usepackage[utf8]{inputenc}
\usepackage{mathtools}
\makeatletter
\def\blfootnote{\xdef\@thefnmark{}\@footnotetext}
\makeatother

\makeatletter
\def\thfootnote{\xdef\@thefnmark{}\@footnotetext}
\makeatother
\usepackage{booktabs}
\usepackage{multicol}
\usepackage{diagbox}
\usepackage[nolist,nohyperlinks]{acronym}
\usepackage{siunitx}
\usepackage{wrapfig}

\begin{document}
\begin{acronym}[MPC] 
\acro{US}{ultrasound}
\acro{DAS}{delay and sum}
\acro{FPU}{floating point unit}
\acro{DL}{deep learning}
\acro{DNN}{deep neural network}
\acro{MSE}{mean square error}
\acro{DMS}{Delay Multiply and Sum}
\acro{MVB}{minimum variance beamforming}
\acro{MV}{minimum variance}
\acro{CFM}{clinical feasablity metric}
\acro{DF}{deepFormer}
\acro{FCNN}{fully convolutional neural network}
\acro{CNN}{convolutional neural network}
\acro{PSNR}{peak signal-to-noise-ratio}
\acro{SSIM}{structural similarity index}
\acro{MS-SSIM}{multi-scale structural similarity index}
\acro{TV}{total variation}
\acro{MSE}{mean square error}
\acro{CNR}{contrast-to-noise ratio}
\acro{SNR}{signal-to-noise ratio}
\acro{SINR}{signal-to-interference-plus-noise ratio}
\acro{FWHM}{full width half maximum}
\end{acronym}
\title{End-to-End Learning-Based Ultrasound Reconstruction}
%
%

\author{Walter~Simson*\inst{1} \and 
R\"udiger~G\"obl*\inst{1} \and 
Magdalini~Paschali\inst{1} \and 
Markus~Kr\"onke\inst{2} \and 
Klemens~Scheidhauer\inst{2}\and
Wolfgang~Weber\inst{2}\and 
Nassir~Navab\inst{1,3}
}
\authorrunning{W. Simson et al.}
%
\institute{Technical University of Munich
\and
Nuclear Medicine, Klinikum rechts der Isar, Munich, Germany
\and
Johns Hopkins University, Baltimore, USA}
\maketitle              
\begin{abstract}
Ultrasound imaging is caught between the quest for the highest image quality, and the necessity for clinical usability.
Our contribution is two-fold: First, we propose a novel fully convolutional neural network for ultrasound reconstruction.
Second, a custom loss function tailored to the modality is employed for end-to-end training of the network.
We demonstrate that training a network to map time-delayed raw data to a minimum variance ground truth offers performance increases in a clinical environment.
In doing so, a path is explored towards improved clinically viable ultrasound reconstruction.
The proposed method displays both promising image reconstruction quality and acquisition frequency when integrated for live ultrasound scanning.
A clinical evaluation is conducted to verify the diagnostic usefulness of the proposed method in a clinical setting.


\end{abstract}
\section{Introduction}

With the ubiquitization of \ac{DL} methods in the medical community, statistical models learned by \acp{DNN} to improve medical imaging have received significant attention.
Specifically in \ac{US} imaging, strides have been made in using \acp{DNN} to interpolate sub-sampled raw data in the case of MLT (a method used in cardiac \ac{US})~\cite{Vedula2018MLT}, improve the filtering of received raw signals~\cite{LuchiesB2018Deep}, and reconstruct \ac{US} filtered images from sub-sampled raw data~\cite{simson2018sub-sampled}.
Lastly, \cite{Nair2018BFAlternative} propose visualizing task-specific data from raw-signals in a binary contrast ``alternative to beamforming" case.
The implicit aim of all these approaches is to extract more information about the tissue being scanned from raw data than can currently be done from reconstructed ultrasound images.
Though these approaches seem promising, to the best of our knowledge, until now there has been no \ac{DNN} proposed to generate clinically viable full contrast ultrasound images---in terms of image quality and frame-rate---end-to-end from raw data by a \ac{DNN}. 

The reconstruction method most used in clinical \ac{US} machines is \ac{DAS}, due to its low computational complexity and data-independence.
On the other hand, data-dependent reconstruction algorithms such as \ac{MV} beamforming, can reconstruct ultrasound images of measurably higher quality, but are still too slow for routine clinical practice.

This dichotomy leads to the following problem: reconstruction algorithms with impressive quantitative results are prohibitively computationally intensive (\ie low-frequency), whereas fast reconstruction algorithms are limited in image quality.

\textbf{Contribution: }We aim at bridging this gap; proposing a novel neural network based on a medical ultrasound specific loss function to reconstruct high quality and clinically relevant images.
We propose an end-to-end learning-based data-dependent ultrasound reconstruction method for real-time applications. 

For the general reader, the following sections briefly discuss the fundamentals of ultrasound imaging that motivate our method.
Ultrasound experts can feel free to proceed to Section~\ref{ssec:nn}.
\begin{figure}[t]
    \centering{   
    \includegraphics[width=\textwidth]{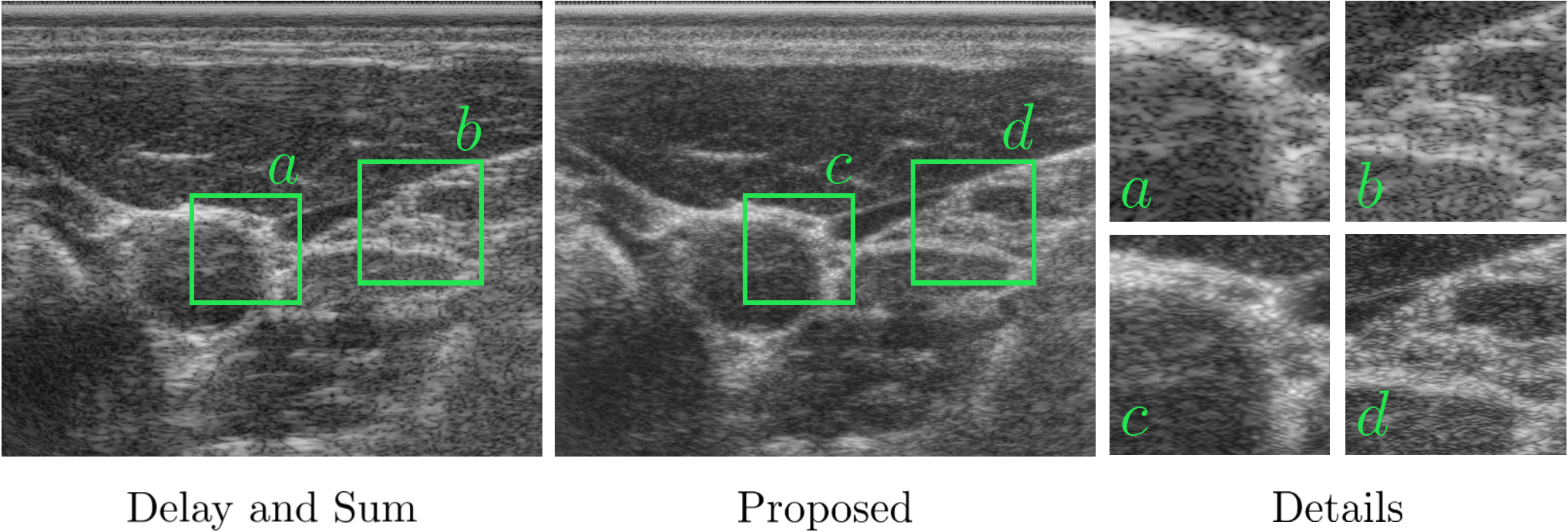}
    \caption{Cross sectional scan of carotid artery. (left) \ac{DAS} reconstruction (middle) the proposed reconstruction method. (right) Detail views. }
    \label{fig:overview_comparison}
    }
\end{figure}

\begin{subsection}{Ultrasound Raw Data}\label{ssec:sli}
Scanline ultrasound imaging describes the process of transmitting and receiving focused ultrasonic beams from a number of ultrasound elements of a transducer. 
Its widespread use for anatomical imaging motivates learned receive beamforming for scanline imaging protocols.
Raw data depicts the voltage measurements caused by the offsets of the piezo elements generated by the pressure wave-front in the tissue.
After reception, the appropriate delays are applied to the individual channels to achieve dynamic receive focusing.
These time-delayed raw data serve as the input for all of following reconstruction methods.
\end{subsection}

\begin{subsection}{Delay and Sum}\label{ssec:das}
The most commonly used reconstruction method today is \ac{DAS}.
After dynamic receive focusing, the signals are multiplied with a constant weight vector (apodization), and summed together to get a local, depth-based signal intensity value.
The signal can be mathematically formulated as:
\begin{equation}
    s_k(n) =  \omega^{T}y_{k}(n)
\end{equation}
where the signal of the $k_{\text{th}}$ scanline $s_k(n)$ is constructed by the multiplication of the apodization window vector $\omega$ with the raw signal $y_{k}$\cite{Szasz2017ABTUS}.
\end{subsection}

\begin{subsection}{Minimum Variance}\label{ssec:mv}
Unlike \ac{DAS}, \ac{MV} beamforming is a data-dependent beamformer (also known as adaptive beamformer) that computes the apodization weights $\omega$ based on statistics of the raw data in order to increase the \ac{SNR} of the output image.
This approach can be described as a maximization problem of \ac{SINR} (\cf Eq.~\ref{eq:maxsinr}), where $R_k$ is the interference-plus-noise covariance matrix, $\sigma_{s}^{2}$ is the signal power and $a$ the steering vector~\cite{Szasz2017ABTUS}.
\begin{equation}
\text{SINR}=\frac{\sigma_{s}^{2}|\omega^{T}a|^{2}}{\omega^{T}R_{k}\omega},
\label{eq:maxsinr}
\end{equation}
\begin{equation}
\underset{\omega}{\text{argmin}}~\omega^T R_k \omega,~\text{ where}~\omega^{T}\textbf{1}=1
\label{eq:minden}
\end{equation}
Since in real-world situations the signal power $\sigma_{s}^{2}$ is unknown, this problem is reformulated as a minimization problem of the denominator described in Eq.~\ref{eq:minden}.
Since this minimization problem must be solved for every reconstructed pixel, it has a high computational cost.

%
%
%

Our highly-parallelized \ac{MV} beamformer achieves a reconstruction frequency of \SI{0.14}{Hz} on an NVIDIA~Titan~V GPU. 
This low frame-rate motivates further reconstruction approaches, that are able to achieve real-time high-quality reconstructions. 
\acp{DNN} offer a potential solution to bridge this gap through a mapping from the space of raw ultrasound signals to the image space. 
\end{subsection}

\section{Method}
We propose a new method of image reconstruction that allows near real-time reconstruction of ultrasound images while offering qualities of traditional state of the art methods.

\begin{subsection}{Learning Beamforming}\label{ssec:nn}
Training is performed with time-delayed raw data as a feature input and \ac{MV} beamformed data as regression target.
To address the challenging task of increasing the clinical efficacy of high-quality ultrasound reconstruction, a neural network is designed and trained to map raw data to image data.

\noindent\textbf{Training Data}
As described in Sec.~\ref{ssec:sli}, time-delayed data is used as an input to the reconstruction.
Time-delayed data ensures the spatial coherency of the 3-D raw input data, which is encoded to the channel dimension of the input.
The output target of the model is \ac{MV}-beamformed data.
Scan conversion  is performed as a discrete subsequent step to ensure generalizability of the method.

\noindent\textbf{Model Architecture}
The proposed network is characterized by: (1) \ac{FCNN}, (2) Encoder-decoder structure, (3) shallow depth, and (4) long- and short-term skip connections.
The selection of a \ac{FCNN} model allows for input size independence, which is required in ultrasound depending on the acquisition protocol and depth.
Moreover, \acp{FCNN} offer accelerated computation due to the lack of their computationally intensive fully connected layers.
Lastly, \acp{FCNN} map spatial relationships of the input data to output features, similarly to the process of \ac{MV}-beamforming.

The encoder-decoder architecture ensures the highest informational density in the bottleneck layer.
The network is kept shallow and only uses three convolutional blocks in the encoder and three in the decoder in order to optimize computation.
Both the shallow network and batch normalization prevent overfitting.
Lastly, the short-term skip connections within the dense convolutional blocks increase reconstruction accuracy and improve gradient flow.
The long-term skip connection from the input to output passes on the fine-grained features from the raw data, such as speckle to the reconstructed image.
In order to prevent data discontinuities on the edges of the data in the forward pass, an occurrence that is not found in ultrasound wave propagation, reflection padding is applied to the input raw data.

\noindent\textbf{Objective Function}
The training loss functions combines the \ac{PSNR} loss, and the \ac{MS-SSIM} loss via a weighting factor $\alpha$.
The \ac{PSNR} loss $\mathcal{L}_{\text{PSNR}}$ is based on the metric of the same name, which is widely used in the field of ultrasound and signal processing. 
It is accompanied by the \ac{MS-SSIM} loss $\mathcal{L}_{\text{MS-SSIM}}$ which serves as a measure for the perceived image reproduction\cite{Zhao2018recloss}.
The combined loss $\mathcal{L}_{\text{US}}$ can then be formulated as:
\begin{multicols}{2}
\begin{equation}
\mathcal{L}_{\text{PSNR}} =  1 - \frac{10 \log_{10}\left(\frac{1}{\text{MSE}(x, y)}\right)}{\text{PSNR}_{\text{max}}}
\end{equation}\break
\begin{equation}
\mathcal{L}_{\text{MS-SSIM}} = 1 - l_M \cdotp \prod_{j=1}^{M}cs_j
\end{equation}
\end{multicols}
\begin{equation}
\mathcal{L}_{\text{US}}=\alpha\mathcal{L}_{\text{MS-SSIM}} + (1-\alpha)\mathcal{L}_{\text{PSNR}}
\label{eq:loss}
\end{equation}
where $\alpha$ is determined to be $0.75$ through empirical experimentation.
\end{subsection}

\begin{subsection}{Experimental setup}\label{ssec:setup}
Reconstruction training is performed on an \textit{in-vivo} data set acquired from five healthy volunteers, ages 23 - 59 (mean 32.4).
In total, 3309 frames are used in the training and validation set with a four to one split between volunteers and five-fold cross-validation.
A test set of two additional volunteers is used for final evaluation (1229 frames). 
The raw input data is cast from the discrete int16 values to float normalized between $0$ and $1$ to improve training dynamics.
Training is performed on an NVIDIA~Titan~V~GPU and all models are implemented in PyTorch.
Training is performed with time-delayed raw data as a feature input and \ac{MV} beamformed data as regression target, for 50 epochs with a learning rate of $10^{-5}$.

    
The inference run-time evaluation was performed on an NVIDIA~Titan~V~GPU as well.
The network is made compatible with libtorch for live inference.
After training, the network weights and input data are cast to half-precision floats, along with the input data, to make use of the higher performance half-precision floating point units on NVIDIA GPUs.
Inference is performed with the C++ interface of libtorch version~1.0.1.
\end{subsection}
\section{Evaluation}
Traditionally, ultrasound image quality is quantified by determining lateral and axial resolution, \ac{CNR}, and the \ac{FWHM}.
These measures are often used when evaluating new ultrasound reconstruction methods.
Unfortunately, these metrics only take visualization of the reconstructed image into account, while disregarding the imaging frequency.
Specifically, beamforming approaches such as \ac{MV} are considered the gold standard in terms of image quality. However, they also suffer from impractical run-times preventing their use in clinical settings.
\subsection{Experimental Evaluation}
We compare our method to the state of the art, w.r.t. the selected loss function, traditional image quality metrics, achieved frame-rate, and clinical acceptance.

\begin{table}[tb]
\centering
\caption{Reconstruction cross-validation comparison using different loss functions.}
\begin{tabular}{r|cc}
Loss Function & SSIM & PSNR \\ \hline
L1 & 0.734 ± 0.0105 & 24.3 ± 0.610\\ 
PSNR          & 0.743 ± 0.0128 & 24.2 ± 0.790 \\ 
\textbf{PSNR-MSSSIM}    & \textbf{0.749 ± 0.009}   & 24.2 ± 0.487 \\ 
\end{tabular}
    \label{tab:quantitative_comparison}
\end{table}

\begin{figure}[tb]
    \centering{   
    \includegraphics[width=0.95\textwidth]{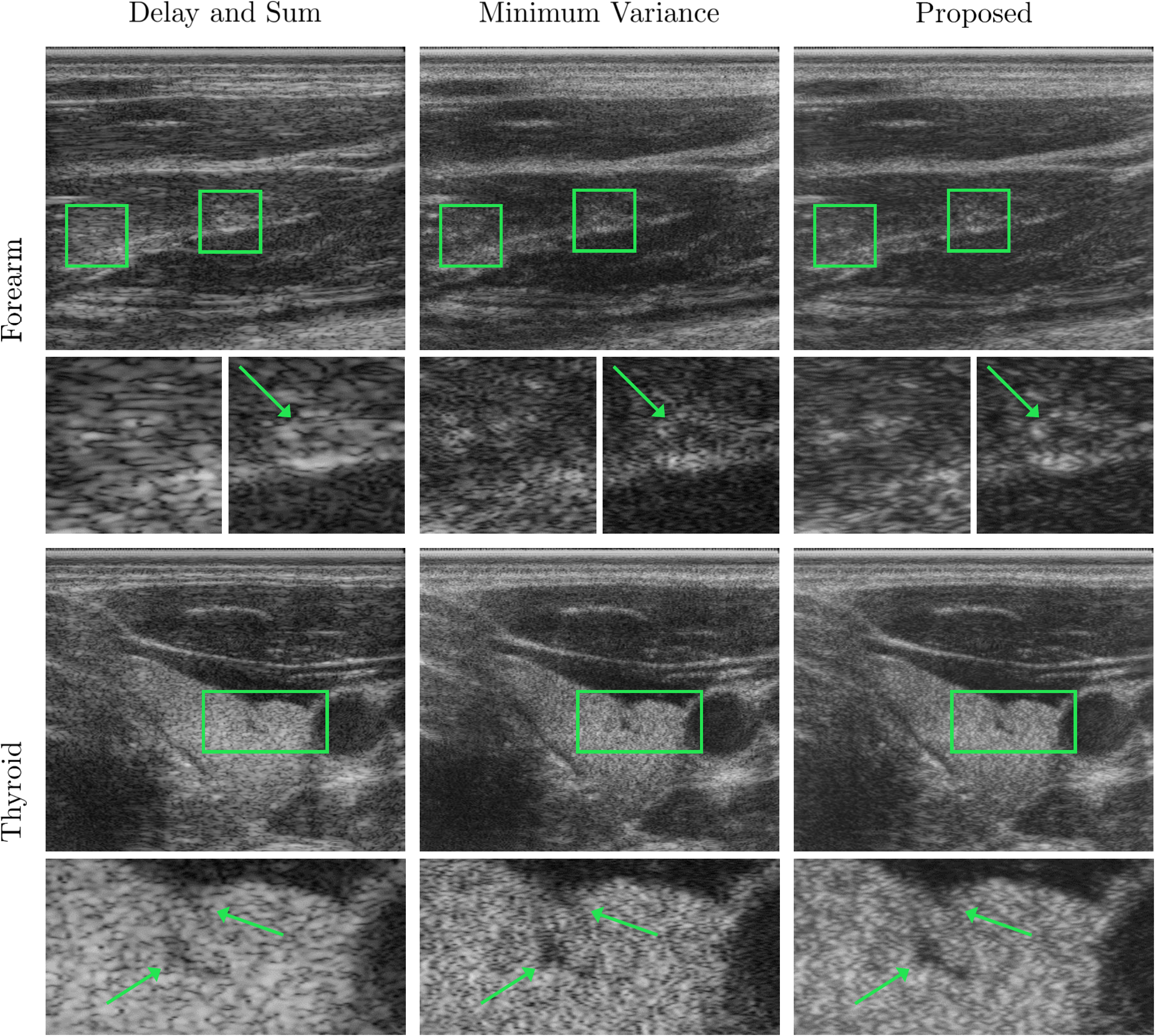}
    \caption{Qualitative evaluation on two anatomies of unseen volunteers. (left) Delay and Sum, (center) Minimum variance, (right) our reconstruction. (top) supinated forearm, interior muscle (Brachioradialis), (bottom) thyroid cross-section.}
    \label{fig:qualitative_comparison}
    }
\end{figure}

A comparative evaluation of the proposed loss can be found in Table~\ref{tab:quantitative_comparison}.
The PSNR-MSSSIM loss performed best w.r.t. both mean and standard deviation of SSIM, exhibiting comparable PSNR.
Consequently, we perform inference to generate evaluation images with the PSNR-MSSSIM loss.
Figure~\ref{fig:qualitative_comparison} shows a qualitative comparison between reconstructions with \ac{DAS}, \ac{MV}, and our method.
In the detail views, arrows point to regions where the reconstruction of fine anatomical structures is compared.
The proposed reconstruction displays the least noise contamination around the indicated annular structure (right detail forearm).
The same observation can also be made in the global view of the forearm scan.
In \ac{MV} and our method, the border of the thyroid (lower detail images), as well as the vessel contained therein, are depicted more clearly.

\begin{table}[tb]
\centering

\caption{Classic ultrasound image quality measures determined using a CIRS 040GSE phantom. Our method was neither trained nor validated on phantom data.}
\begin{tabular}{r|ccccc}
     & {CNR [\SI{}{dB}]} & \multicolumn{2}{c}{FHWM [\SI{}{mm}]}          & \multicolumn{2}{c}{Resolution [\SI{}{mm}]}    \\ \hline
                        &                      & Axial          & Lateral         & Axial        & Lateral      \\ 
DAS                     & 10.7                 & 0.339          & 0.780           & 0.25         & 1.0          \\ 
MV                      & 9.30                 & 0.326          & 0.301           & 0.25         & 1.0          \\ 
Ours                    & 8.68                 & 0.332          & 0.425           & 0.44         & 1.0     
\end{tabular}
\label{tab:quality_assessment_results}
\end{table}
For a direct quantitative comparison, we evaluate our method on wire and cyst phantoms (\cf Table~\ref{tab:quality_assessment_results}).
Though all phantoms are unseen in both train and test sets, the proposed method performs comparably to \ac{MV}.
To investigate the imaging frequency of all considered algorithms, we perform live acquisitions and monitor the frame-rates.
The resulting acquisition frequencies are \SI{21}{Hz}, \SI{0.14}{Hz}, and \SI{17}{Hz} for \ac{DAS}, \ac{MV} and our reconstruction respectively.


\subsection{Clinical Evaluation}
\begin{wraptable}{r}{5.5cm}
\centering
\caption{Results of clinical evaluation on a five-point scale}
\begin{tabular}{r|ccc}
Metric    & MV       & DAS     & Ours   \\ \hline
Contrast & 3.26 & 2.82 & \textbf{3.35} \\
Texture & 3.13 & 1.93 & \textbf{3.14} \\
Speckle & \textbf{2.89} & 1.78 & \textbf{2.89} \\
Resolution & 2.88 & 2.15 & \textbf{2.95} \\
Gross anatomy & 3.14 & 2.63 & \textbf{3.18} \\
Fine anatomy & 2.32 & 1.79 & \textbf{2.47} \\
Artifact min. & \textbf{2.73} & 2.26 & 2.67 \\ \hline
Average & 2.90 & 2.19 & \textbf{2.95}
\end{tabular}
\label{tab:qualitative_eval}
\end{wraptable}

Raw data acquisitions of the thyroid glands of volunteers are conducted by a trained nuclear medicine physician for clinical evaluation purposes.
These include the gross anatomy and detailed views of thyroid nodules, both in transverse and axial views.
Nodules are present in all three volunteers.
Though our method is capable of real-time use, the \ac{DAS} beamformer is used during all acquisitions in order to avoid bias.

Using \ac{DAS}, selected frames with clinically relevant information are chosen offline by the acquisition physician.
These frames are subsequently reconstructed with all considered methods and evaluated by four independent physicians; two male, two female, with 1, 3, 3, and 30\texttt{+} years of clinical experience respectively.

Each physician is provided a series of 24 ultrasound image sets, similar to those in Fig.~\ref{fig:qualitative_comparison}.
Each set consists of three images in random order, all generated from the same raw channel data; one \ac{MV}, one \ac{DAS} and one image generated with our proposed method.

The physicians evaluate each of the images on seven qualitative metrics; contrast, image resolution, anatomy texture, overall image speckle, anatomy visibility (gross and fine), and the presence of artifacts in the image.
All metrics are rated on a five-point scale.
Table~\ref{tab:qualitative_eval} shows the average rating per metric and reconstruction method.
The proposed method was rated highest in all categories, except artifact minimization, where it is only marginally outperformed by \ac{MV}.

\section{Discussion}
Based on the evaluation of acquisition frequencies, we see that the proposed method with a frame-rate of \SI{17}{Hz} offers a clinically viable reconstruction method.
The qualitative results (\cf~Fig.~\ref{fig:qualitative_comparison}) display a very promising visualization of fine anatomical structures in the reconstruction.
This initial sign of promising \textit{in-vivo} reconstruction quality is affirmed by the four physician evaluation.

Performance discrepancies on the phantom data (\cf Table~\ref{tab:quality_assessment_results}) can most likely be attributed to the fact that the network is only trained on \textit{in-vivo} data.
Augmenting the training set with non-organic data samples could prove to allow for more representative image quality metric evaluation.
Similar to data-dependent reconstructions, the perceptive field of the proposed method allows for ``decisions" about local signal intensities to be derived from surrounding signal information.
Thanks to the fully convolutional nature of the network, our method can be applied to images of arbitrary depths.
Furthermore, properties of the images reconstructed by a \ac{DNN} can be ``steered" via the selection of an appropriate loss function and loss function regularizing term.

Lastly, the reported frame rate numbers show the immense potential for processing pipelines utilizing the advantages of neural networks to model complex behavior in competitive run-times.
When the quality of \ac{MV} and the proposed method are considered to be comparable, the proposed method offers a total speed-up of $\frac{\text{f}_{\text{Ours}}}{\text{f}_{\text{MV}}} = \frac{\SI{17}{Hz}}{\SI{0.14}{Hz}} = 121.$

\section{Conclusion}
Deep neural networks offer a promising alternative to high quality data-dependent beamforming methods.
We have shown that gold standard approaches such as \ac{MV} can be modeled by a \ac{DNN} while offering substantial speed-up.
Our clinical evaluation shows that physicians find the appearance of the proposed method visually pleasing and feel they can better extract clinically relevant information (\cf~Table~\ref{tab:quality_assessment_results}).
In this work, a new ultrasound reconstruction method is presented as an opportunity to bridge the gap between real-time beamformers and high-quality data-dependent beamformers.
A suitable network architecture is defined and validated in both experimental and clinical contexts.
By defining an ultrasound specific loss function to be used during training, images created with the proposed method can be steered towards relevant clinical objectives.
We demonstrate the clinical feasibility of such an approach by integrating the trained network in an ultrasound device for real-time inference and \textit{in-vivo} evaluation.


%
%
%
\bibliography{DeepFormer}
\bibliographystyle{splncs04}
\end{document}